# A New Similarity measure for taxonomy based on edge counting

Manjula Shenoy.K[1], Dr.K.C.Shet2, Dr. U.Dinesh Acharya[1]

[1]Department of Computer Engineering, Manipal University, MIT, Manipal
`manju.shenoy@manipal.edu, dinesh.acharya@manipal.edu`
[2]Department of Computer Engineering, NITK, Suratkal
`kcshet@rediffmail.com`

## ABSTRACT

This paper introduces a new similarity measure based on edge counting in a taxonomy like WorldNet or Ontology. Measurement of similarity between text segments or concepts is very useful for many applications like information retrieval, ontology matching, text mining, and question answering and so on. Several measures have been developed for measuring similarity between two concepts: out of these we see that the measure given by Wu and Palmer [1] is simple, and gives good performance. Our measure is based on their measure but strengthens it. Wu and Palmer [1] measure has a disadvantage that it does not consider how far the concepts are semantically. In our measure we include the shortest path between the concepts and the depth of whole taxonomy together with the distances used in Wu and Palmer [1]. Also the measure has following disadvantage i.e. in some situations, the similarity of two elements of an IS-A ontology contained in the neighbourhood exceeds the similarity value of two elements contained in the same hierarchy. Our measure introduces a penalization factor for this case based upon shortest length between the concepts and depth of whole taxonomy.

## 1. INTRODUCTION

The question of similarity identification and/or the computation of semantic distances are regarded as a research subject highly investigated in the fields of data processing, Artificial Intelligence, and Linguistics. In particular, the field of the information retrieval which is largely based on the similarity identification measures between documents. The problem of those approaches is that they typically focus on the single words of a document ignoring the ontological relationships that exist between the words. We can distinguish three ways to determine the semantic similarity between objects in ontology. The first approach indicates the evaluation of the similarity by the information content (also called the *node based* approach). The second approach represents an evaluation of the similarity based on conceptual distance (also called *edge based* approach). The third approach is hybrid which combines the first two approaches. The problem of the second approach is dependent on the ontology construction. Furthermore, this approach, adopting IS-A ontology, presents the following disadvantage: in some situations, we can obtain a similarity value of two elements of an ontology contained in the neighborhood which exceeds the value of similarity of two concepts contained in the same hierarchy. This situation is inadequate within the information retrieval framework. Also the shortest distance between the two needs to be considered while calculating similarity measure in applications like ontology matching. In order to overcome this problem, we propose, in this Paper, a new similarity measure giving realistic results and closer relations to reality for concepts not located in the same path. The paper presents a similarity measure that is suited for Comparing concepts in ontology. Although finding similar concepts is a core task in the area of ontology alignment/merging [5] [6]. The proposed measure can be adopted effectively in this field.

The remainder of this paper is organized as follows: Section 2 reviews the literature on similarity measures. Section 3 simulates similarity measure to the conceptual proximity. Section 4 is a detailed presentation of our similarity measure with some examples. The experimental results of our prototype and a comparison with other works are presented in section 5. Finally, section 6 concludes with some future perspectives.

## 2. RELATED WORK

We can distinguish three main approaches for the similarity identification measures between the taxonomy objects. The first type is based on the nodes [2] [7] [8]. Works under the banner of these approaches used the typically information based content to determine the conceptual similarity. Moreover, the similarity between two concepts is obtained by the degree of sharing information. The second type is based only on the hierarchy or the edge distances [1] [9] [10] [11]. The problem with this approach is that the taxonomy arcs represent uniform distances, i.e. all the semantic links have the same weight. Finally, the hybrid approach [12] [13] [14] [15] which combines the two approaches presented above. With these approaches, there exist several manners of detecting conceptual similarity of two words in a hierarchical semantic network. The following section presents some measures which are listed under these approaches.

### 2.1. Wu and Palmer Measure

The principle of similarity computation is based on the edge counting method which is defined as follows: Given an ontology formed by a set of nodes and a root node (R) (Fig. 1). C1 and C2 represent two ontology elements of which we will calculate the similarity. The principle of similarity computation is based on the distance (N1 and N2) which separates nodes C1 and C2 from the root node and the distance (N) which separates the closest common ancestor (CS) of C1 and C2 from the node R. The similarity measure of Wu and Palmer [1] is defined by the following expression:

$$Sim_{WP}=2*N/(N1+N2)$$

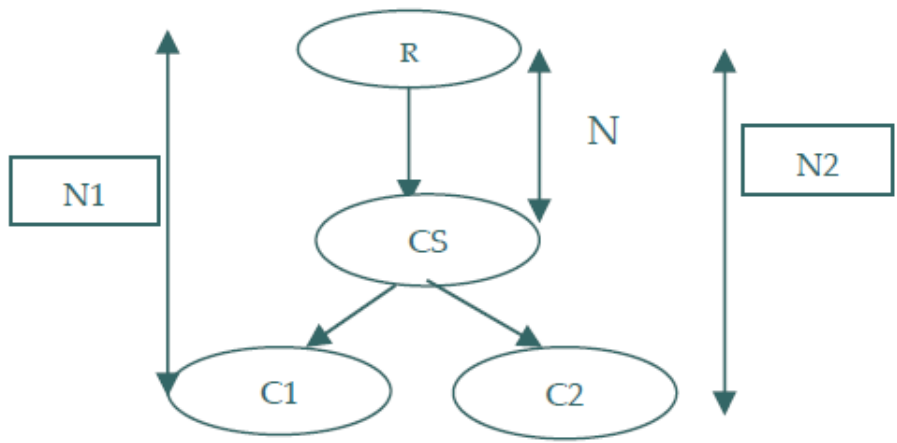

Figure 1. Example of a concept hierarchy

The problem resulting from this measure is that the arcs in ontology represent uniform distances (i.e. all the semantic links have the same weight). A comparison between the methods of similarity measure is carried out by [2]. This comparison reveals that the Wu and Palmer measure [1] has the advantage of being simple to calculate, in addition to the performances which it presents while remaining as expressive as the others. For this reason we have adopted this measure as a base for our work.

## 2.2. Rada et al. Measure

This measure [10] is adopted in a semantic network and it is founded on the fashion that we can calculate the similarity based on the hierarchical links "IS-A". To calculate the similarity of two concepts in ontology, we must calculate the number of the minimal arcs which separate them. This measure, based on the edge counting between nodes by the shortest way, presents a mean of the most obvious to evaluate the semantic similarity in a hierarchical ontology.

## 2.3. Ehrig et al. Measure

A work of similarity measure based on ontology was introduced by [11]. This work presents three layers: data, ontology and context. The similarity of the entities is measured on the data level by considering the data values of simple or complex types (integer, strings). The semantic relationships between the entities are measured on the level of ontology layer. The context layer specifies how the ontology entities are used in a certain external context, more specifically, the application context. All the previously listed similarities are calculated as function amalgams which combine the similarity measure of the individual layers.

## 2.4. Resnik Measure

This was [7] [14] the first to bring together lexical database and corpora, and calculates similarity by considering the information content (IC) of the LCS of two concepts, expressed by:

$$Sim_{res}(c_1, c_2) = IC(lsc(c_1, c_2))$$

## 2.5. Lin's Measure

Lin's [21] similarity measure follows from his theory of similarity between arbitrary objects. The *lin* measure augment the information content of the LCS with the sum of the information content of concepts c1 and c2 themselves. The *lin* measure scales the information content of the LCS by this sum:

$$Sim_{lin}(c_1, c_2) = \frac{2 \times IC(lcs(c_1, c_2))}{IC(c_1) + IC(c_2)}$$

## 2.6. Jiang and Conrath Measure

This measure is based on a combination of using edge counts in the WordNet is-a hierarchy and using the information content values of the WordNet concepts, as described in the paper by Jiang and Conrath [12]. Thus the information content of the two nodes, as well as that of their most specific sub sumer, plays a part. Different from the *lin* measure, jcn takes the difference of this sum and the information content of the LCS:

$$Sim_{jcn}(c_1, c_2) = \frac{1}{IC(c_1) + IC(c_2) - 2 \times IC(lcs(c_1, c_2))}$$

**2.7. Leacock and Chodorow measure**

Leacock and Chodorow [13] rely on the length len(c1,c2) of the shortest path between two Synsets for their measure of similarity. However, they limit their attention to is‑a links and scale the path length by the overall depth D of the taxonomy:

$$Sim_{lch}(c_1, c_2) = \log\left(\frac{2 \times D}{len(c_1, c_2)}\right)$$

## 3. SIMILARITY MEASURE AND SIMULATION OF CONCEPUAL SIMILARITY

In Fig. 2, we present a graph representing a hierarchy of the concept. This graph represents an ontology extract of pedagogic field.

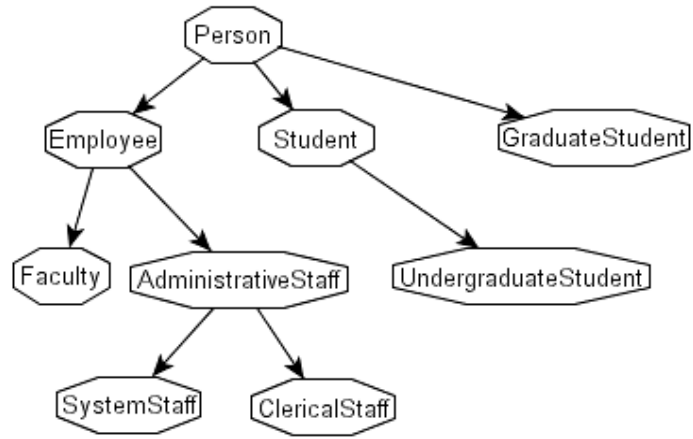

Figure 2. Concept hierarchy graph

The concepts contained in this ontology represent intuitively a set of the varied conceptual distances if they are compared between them. As an example the two concepts "Student" and "GraduateStudent" presents similarity value equal to 0 in the case of the use of traditional similarity measure, which includes external information with the hierarchy such as measure of [15][16]. On the other hand, the adoption of an approach based on the hierarchy gives a similarity measure different from 0 for these two same concepts. Moreover, the similarity value of two concepts "Student" and "GraduateStudent" are lower than the concepts "Student" and "UndergraduateStudent". However, we judge that the concept "Student" is closer with the concept "UndergraduateStudent"than the concept "Graduate Student". These precise details are very interesting for the research of the semantic similarities of concepts set contained in ontology. These intuitive distances can be used, for example, to the improvement of the search engines on the level of the effectiveness and precision of answers to the user's requests. The simplest structure supporting the reasoning on the hierarchy of types is that which can be found in a support of conceptual graphs. In this structure, the IS-A links group the types according to the definitional characteristics which they share. The arrows presented in Fig. 3 present the relation IS-A from a superclass to its subclasses.

# 4. FORMULA AND MODEL OF PROPOSED SIMILARITY MEASURE.

## 4.1. Ontology Formalism

Let be an ontology which is a finite set of classes and seems to be equated with a rooted tree. We denote by (C, P,HP, HC) the elements of where C and P indicate, respectively, the set of classes and the set of properties contained in . The hierarchies HP and HC indicate, respectively, the hierarchy of properties and the hierarchy of classes of . The measure of [1] is interesting but presents a limit because it primarily aims to detect the similarity between two concepts compared to the distance of their least common subsumer. The more this subsuming is general, the less similar they are (and conversely). However, it does not collect the same similarity as the symbolic conceptual similarity (conSim). Thus we can obtain Simwp (A, D) < conSim (A, B),D being one descendant of A and B one of the brothers of A. This situation is inadequate within the information retrieval framework where it is necessary to turn up all descendants of a concept (i.e. request) before its vicinity. For example, we can obtain with this measure, a value of similarity between the concept "PostDoc" and "AdministratifStaff" which exceeds the value of similarity between "Person" and "PostDoc". However, this measure offers a higher similarity between a concept and its vicinity compared to this same concept and a concept contained in the same path (see example 1).

Example 1: Let the ontology of figure 3, we indicate by C1, C2 and C3 the concepts "Person", "PostDoc" and "AdministrativeStaff". Simwp (C1, C2) =2*1/ (1+4) =0.4 and
Simwp (C2, C3) =2*2/(4+3)=4/7=0.57.

## 4.2. Measure Formula

We put forward a new measure which compensates for the previous problem. The formula is
Sim= $(2N.e^{-L/D})/N1+N2$.

Here L is the shortest distance between the two concepts computed as follows. i.e. for every edge passed in the vertical direction we give a weight of 1 and when we change the direction we give that edge a weight of 1 more. For example in Fig 3. The l for PostDoc and Administrative staff is 4 and PostDoc and Person are 3. D is the depth of whole ontology tree. N1 and N2 are the distances from root node to Concept 1 and Concept2 like Wu and Palmer measure.   is 1 for neighbourhood concepts and 0 for concepts from same hierarchy.

## 4.3. Property of proposed measure

In this section we enumerate some properties of similarity measure [17]. These properties depend on a particular application; sometimes a property will be useful, sometimes it will be undesirable. The function of similarity which we propose ensures the following properties: Being given three concepts A, B and C of ontology:

1) Nonnegativity: Sim (A, B)  0,
2) Identity: Sim (A, A) = Sim(B, B) =1;
3) Symmetry: Sim(A, B) = Sim(B, A);
4) Uniqueness: Sim (A, B) =1 implies A=B;

### 4.4. Relevance of similarity measure

In our context, a similarity measure is relevant, if it presents a value for each couple of concepts (A, Bi) contained in the same hierarchy, which is always higher or equal to this same concepts and any neighboring concept (A,Ci). i.e. ∀ concept Bi descendant of A and ∀ concept Ci neighbors of A, there exist Sim (A, Bi) ≥ Sim (A, Ci).

## 5. EXPERIMENTAL RESULTS

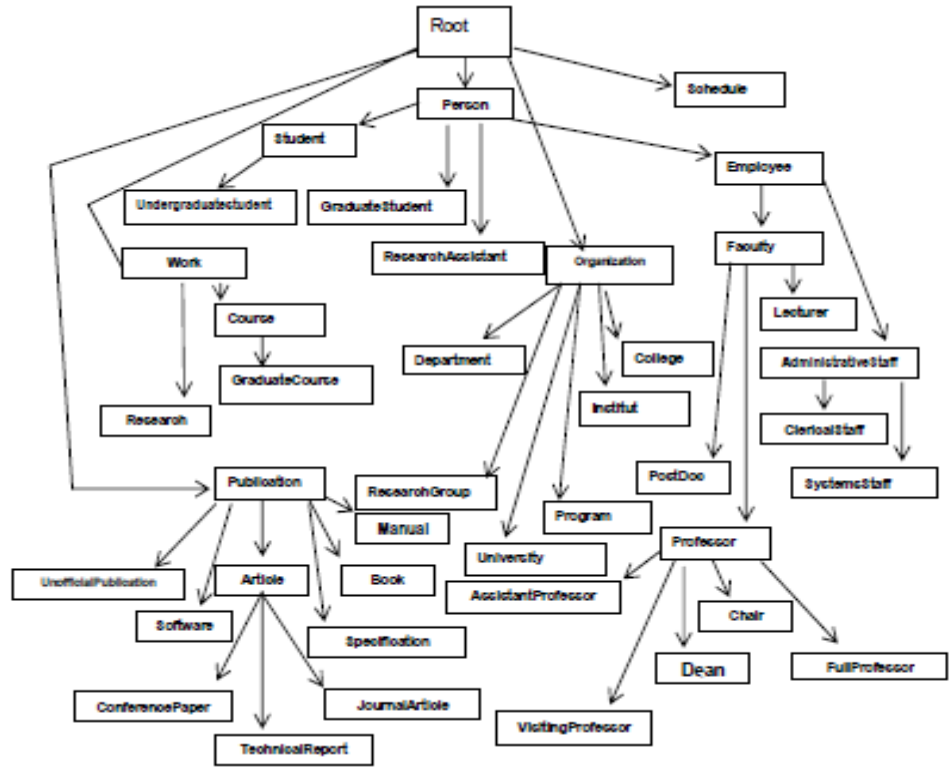

Figure 3. Univ-bench ontology

Table I gives Wu and Palmer[1] and our similarity measure for the concepts in univ-bench ontology shown in Fig.3. Our measure is advantageous because it leads to a lower similarity value for close concepts compared to concepts in the same hierarchy.

Table I

| C1 C2 | $Sim_{wp}$ | Sim |
|---|---|---|
| Person, ResearchAssistant | 0.66 | 0.66 |
| VisitingProfessor,FullProfessor | 0.8 | 0.8 |
| VisitingProfessor, SystemStaff | 0.44 | 0.26 |
| ResearchAssistant,Faculty | 0.4 | 0.29 |
| Chair,AdministrativeStaff | 0.5 | 0.33 |
| Research,GraduateCourse | 0.4 | 0.29 |
| SystemStaff ,Professor | 0.5 | 0.33 |
| Systemstaff,Dean | 0.44 | 0.29 |
| Person Schedule | 0 | 0 |
| Person, Work | 0 | 0 |

| Person,Student | 0.66 | 0.66 |
| --- | --- | --- |
| Student,Undergraduatestudent | 0.8 | 0.8 |
| Student,GraduateStudent | 0.5 | 0.4 |
| Student,Publication | 0 | 0 |

## 6. CONCLUSIONS

In this work we have presented an extension of similarity measure defined by Wu and Palmer. The measure is found to be good. The future work would be to compute similarity between a set of concepts rather than two concepts in the hierarchy.

## Authors

Manjula Shenoy.K is currently working as a Assistant professor (Sel.Grade) at CSE Department, Manipal Institute of Technology, Manipal  University, Manipal

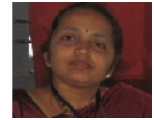

Dr. K.C. Shet is a Professor in Department of Computer Engineering, National Institute of  Technology ,Suratkal. He has several Journal and Conference papers to his credit.

Dr. U.Dinesh Acharya is Professor and Head of the department of Computer Science and Engineering, Manipal Institute of Technology , Manipal University,Manipal. He has several Journal and Conference papers to his credit.
.

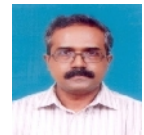